%% file: main.tex
\documentclass[usemulticol,english]{DDCLSconf}
\usepackage[comma,numbers,square,sort&compress]{natbib}
\usepackage{epstopdf}
\usepackage{amsmath}
\usepackage{amssymb}
\usepackage{mathtools}
\usepackage{amsthm}
\usepackage{subcaption}
\usepackage{multirow}
\usepackage{multicol}
\usepackage{url}
\begin{document}

\title{Efficient Fusion and Task Guided Embedding for End-to-end Autonomous Driving}

\author{Yipin Guo,
        Yilin Lang,
        Qinyuan Ren}



\affiliation{College of Control Science and Engineering, Zhejiang University, Hangzhou 310027, P.~R.~China, \email{\{guoyipin, langyilin, renqinyuan\}@zju.edu.cn}}

\maketitle

\begin{abstract}
To address the challenges of sensor fusion and safety risk prediction, contemporary closed-loop autonomous driving neural networks leveraging imitation learning typically require a substantial volume of parameters and computational resources to run neural networks. Given the constrained computational capacities of onboard vehicular computers, we introduce a compact yet potent solution named \textbf{EfficientFuser}. This approach employs EfficientViT for visual information extraction and integrates feature maps via cross attention. Subsequently, it utilizes a decoder-only transformer for the amalgamation of multiple features. For prediction purposes, learnable vectors are embedded as tokens to probe the association between the task and sensor features through attention. Evaluated on the CARLA simulation platform, EfficientFuser demonstrates remarkable efficiency, utilizing merely 37.6\% of the parameters and 8.7\% of the computations compared to the state-of-the-art lightweight method with only 0.4\% lower driving score, and the safety score neared that of the leading safety-enhanced method, showcasing its efficacy and potential for practical deployment in autonomous driving systems.

\end{abstract}


\footnotetext{Best Student Paper Award of the IEEE 13th Data-Driven Control and Learning Systems Conference. This work was supported by NSFC 62088101 Autonomous Intelligent Unmanned Systems.}

\vspace{-2em}

\input{section/1-Introduction}

\input{section/3-Method}
\input{section/4-Experiments}
\input{section/5-Conclusion}

\bibliography{refer}
\bibliographystyle{ieeetr}

\end{document}

%% file: section/1-Introduction.tex
\section{Introduction}
    The success of deep neural networks (DNNs) has paved the way for data driven learning-based approaches to autonomous driving (AD), utilizing large-scale data and computation. This has made end-to-end autonomous driving (E2E AD) systems a reality. Defined as fully differentiable programs that directly map raw sensor data to planned actions or low-level control commands, E2E AD eliminates the need for intermediate modules. This not only simplifies development but also has the potential to improve performance.

    E2E AD primarily advances along two main directions: reinforcement learning \cite{Kendall2019DriveInDay, Toromanoff2020ModelFreeRL} (RL) and imitation learning (IL). While RL robustifies against data distribution shifts, recent advancements in driving scene generation  \cite{shao2023reasonnet}, coupled with the growing availability of data from electric vehicles, have made IL increasingly attractive.

    Early E2E AD with IL \cite{bojarski2016firstEnd2End} utilized convolutional neural networks (CNNs) to extract image features and directly imitate control actions. However, limited by data availability and computational power, these early systems struggled to achieve good performance. Subsequently, most research shifted towards predicting trajectories (i.e. waypoints). LBC \cite{Chen2019LBC} utilizes policy distillation where a teacher model, trained with ground-truth BEV semantic maps, predicts future waypoints. A student model, using only image data, learns from the teacher's predictions. TransFuser \cite{Prakash2021transfuser1,Chitta2023transfuser2} uses two CNNs to extract image and LiDAR information, followed by transformers to fuse these information at each downsampling step. Similarly, InterFuser \cite{shao2022interfuser} extracts sensor features with CNNs, but leverages an encoder-decoder structure to additionally incorporate traffic rules and vehicle density information, aiming for a safe driving strategy. TCP \cite{Wu2022TCP} combines two kinds of prediction goals, introduces control actions and waypoints within a sequence of time into training, and uses a control method that mixes trajectory trackers and behavior prediction results. 

    Despite their promise, end-to-end approaches face a critical hurdle: the mismatch between DNNs' hefty computational demands and AD's need for real-time, low-latency operation. Current on-board hardware struggles to handle the complex architecture and massive parameters of DNNs, limiting the real-world feasibility of end-to-end systems. Recently, more and more works have focused on lightweight neural network (NN) designs. MobileNet \cite{sandler2019mobilenetv2,howard2019mobilenetv3} utilize depthwise separable convolutions for computational efficiency without losing accuracy. MCUNet \cite{lin2020mcunet,lin2021mcunetV2} pushes the limits of miniaturization, fitting networks onto embedded platforms with only 256KB of memory, significantly expanding AI deployment possibilities in resource-constrained environments. Despite advancements in lightweight NN designs, there remains a scarcity of such networks specifically tailored for E2E AD systems.

    To overcome the computational hurdles of E2E AD, we introduce \textbf{EfficientFuser}, a powerful and hardware-friendly model, which uses EfficientViT \cite{Liu2023EfficientVit} for feature extraction from multi-view and employs a decoder-only transformer with task guided embedding for prediction. Cross attention seamlessly integrates multi-view information at different scales. Remarkably, EfficientFuser maintains strong capabilities and efficiency for requiring fewer parameters and computations, making it ideal for applications. The main contributions can be summarized as follows:

    \begin{itemize}
        \item EfficientFuser fuses multiple camera views through cross attention, providing a richer understanding of the environment without straining much computation.
    
        \item A decoder-only transformer is used for the prediction process. With learnable vectors as embedded tokens, the decoder finds the connection between tasks and sensor features through attention.
    
        \item The predicted waypoints and control inputs are mixed dynamically, offering flexible adaptation to diverse driving scenarios and potentially safe behaviors. 
        
    \end{itemize}

%% file: section/3-Method.tex
\begin{figure*}[t]
  \centering
  \includegraphics[width=\hsize]{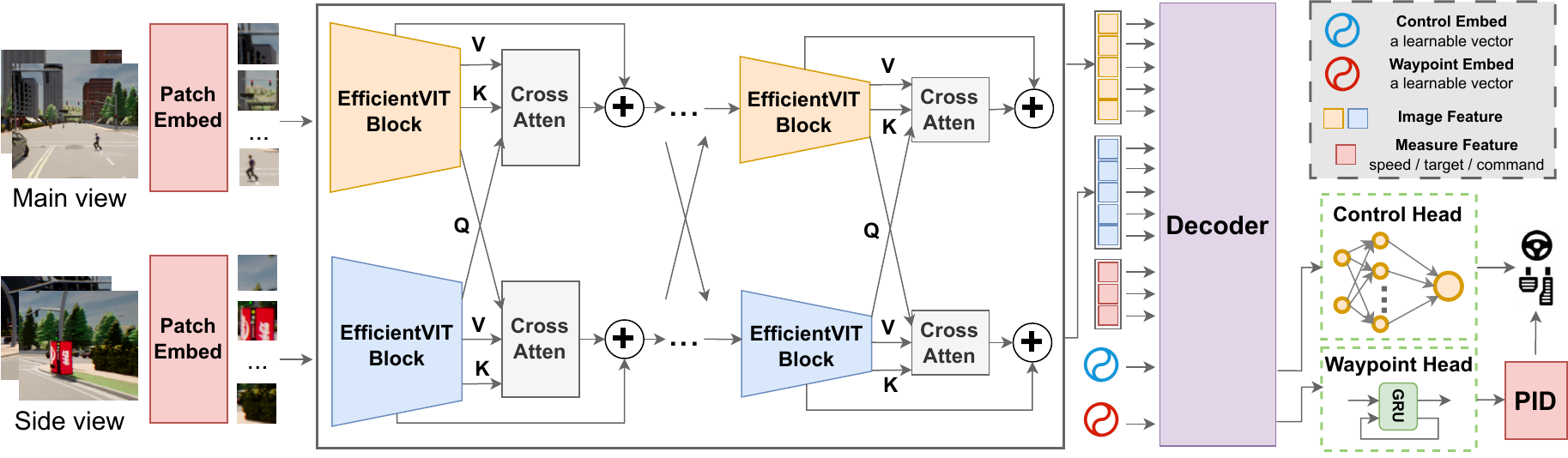}
  \caption{\textbf{Architecture.} \textit{Image feature extraction:} Consider multi-view RGB images as inputs to EfficientFuser which uses several cross attention for the fusion of feature maps between different views. EfficientViT is used as the visual backbone, which makes it suitable for cross attention while maintaining a low computational resource footprint. Since the main-view and focus-view images are more important, larger backbones are used, and side-view images use smaller backbones. \textit{Decoder:} A decoder-only transformer is used to make predictions. Visual features and sensor features are input as tokens. Additionally, two trainable vectors are set up as tokens, learning relevant information with other tokens at the early stage in the decoder layer.}
  \label{fig:main}
  \vspace{-1.5em}
\end{figure*}

\section{EfficientFuser}
    In this work, an architecture for end-to-end driving is proposed, as shown in Fig. \ref{fig:main}, with three components: (1) Cross Attention feature fusion. (2) Decoder-only transformer for prediction. (3) Dynamically mix for behavioral prediction results and waypoint tracker actions. The following sections detail the problem setting, input and output, and each component of the model.
\subsection{Problem Setting}
    EfficientFuser's decision-making process is anchored by an extensive input state, symbolized as $x$, which amalgamates multiple data sources: sensor signal $i$, capturing real-time environmental conditions via vehicle camera; vehicle speed $v$, indicating the current velocity; and high-level navigation information $g$, comprising discrete navigation commands and target coordinates from the global planner. To modulate the vehicle's speed and direction, the system generates outputs for $throttle \in [0, 1]$, $brake \in [0, 1]$ and $steer \in [-1, 1]$, controlling acceleration, deceleration, and steering respectively.

    The goal of IL is to learn a policy $\pi$ that imitates the behavior of an expert $\pi^*$. TCP believes that both trajectory and control actions contain important driving information, so both should be used as imitation targets. EfficientFuser follow the settings in TCP. The difference is that TCP first predicts the trajectory and then uses it to guide the prediction of control actions, while the information between the two can be fully exchanged through attention \cite{Vaswani2017transformer} in the decoder and predicted at the same time for EfficientFuser. Imitation target can be formulated as:
    \begin{equation}
        \label{equ:target}
        \arg \min_{W}  E_{D\sim (x, [\pi_t^*,\pi_c^*])}[\mathcal L(\pi_t,\pi_t^*) + \mathcal L(\pi_c,\pi_c^*)],
    \end{equation}
    where $D\sim (x, [\pi_t^*,\pi_c^*])$ is a dataset comprised of state-action pairs collected from the expert. $\pi_c$ denotes the policy of control branch and $\pi_t$ denotes the trajectory prediction. $\mathcal L$ is the loss measuring how close the action from the expert and the action from the model is.

    Roach \cite{Cho2014Roach} is used as the expert, which is a relatively straightforward model that has been trained using RL with access to privileged information. This information encompasses various aspects of the driving environment such as roads, lanes, routes, vehicles, pedestrians, traffic lights, and stop signs, all of which are rendered into a 2D Bird's Eye View (BEV) image. Compared with experts made by hand-crafted rules, it can provide latent features for control action prediction as intermediate supervision, making the training more stable. 

\subsection{Architecture Design}

    A structure that combines visual transformer (ViT) \cite{dosovitskiy2020vit} and decoder-only transformer in large language model (LLM) \cite{Tom2020GPT, Dong2019UniLM} is used. The transformer has since revolutionized various fields, including computer vision and even time-series modeling, which can be formulated as
    \begin{equation}
        \label{equ:transformer}
        out = Transformer(x) = MLP(Attention(x)).
    \end{equation}
    The core innovation of transformer lies in the attention mechanism, which allows the model to weigh the importance of different parts of the input data. The calculation of attention can be expressed as
    \begin{equation}
        \label{equ:attention}
        out = Attention(Q,K,V) = softmax(\frac{QK^T}{\sqrt{D_k} } )V.
    \end{equation}
    These mechanisms involve mapping the input to three vectors: Query ($Q$), Key ($K$), and Value ($V$), often through linear layers. $D_k$ is the dimension size of $K$, which is scaled to prevent training instability. When dealing with a single input, these vectors all correspond to the same input, known as \textit{self-attention}, allows the model to find inherent connections. When working with two inputs, the $Q$ originates from one source, while the $K$ and $V$ come from the other. This \textit{cross-attention} mechanism enables $Q$ from different source to selectively focus on relevant features.

    \subsubsection{Image Backbone}

    EfficientViT \cite{Liu2023EfficientVit} is used as the visual backbone. ViT pioneered the paradigm shift of applying the transformer architecture, originally designed for natural language processing, to the domain of computer vision. It firstly divides an input image into smaller patches (e.g., 16x16 pixels). These patches are then flattened and treated as a sequence of tokens, similar to words in a sentence. Then self-attention mechanism of the transformer operates on these image patch tokens. This allows ViT to learn complex relationships between different regions of the image.

    While standard ViTs often require significant computational resources, EfficientViT with Cascaded Group Attention can maintain efficiency. This approach allows faster processing and reduced memory usage.


    
    Unlike CNN-based fusion approaches Transfuser, which relies on multiple transformer layers for fusion after each downsampling, ViT can work with small image patches directly. For the Transfuser, CNNs still necessitate feature pooling before the transformer stage to reduce the computational burden, potentially followed by interpolation to restore resolution. This interpolation can compromise the integrity of the original feature map.
    
\begin{figure}[h]
    	\centering
    	\includegraphics[width=0.9\linewidth]{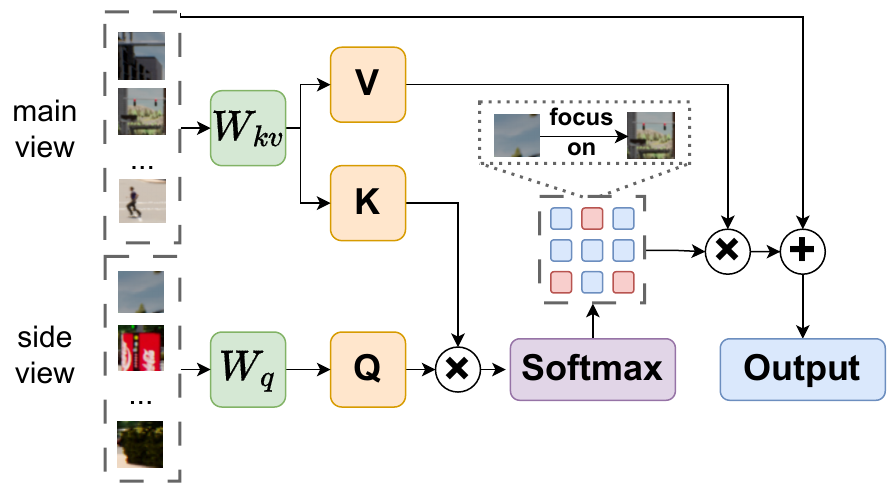}
        \caption{Fusion with Cross Attention.}
        \label{fig:crossAttention}
        \vspace{-1.5em}
    \end{figure}

    To retain the original image features, a method that leverages cross-attention between the two image backbones is employed, which enables knowledge exchange between both perspectives, guiding them to focus on relevant features in specific areas as Fig. \ref{fig:crossAttention}.
    \subsubsection{Decoder Transformer}
    In the realm of sequence processing, transformers often rely on an encoder-decoder structure \cite{Vaswani2017transformer}. The encoder first processes the input sequence (image patches) and extracts essential information, culminating in a context vector that captures the entire input's essence. The decoder then takes over, utilizing the context vector and query from additional information (speed, command, occupation map, etc.) to generate the output sequence element by element. This is how InterFuser works.

    Inspired by the prevalent large language models like GPT \cite{Tom2020GPT}, we employ a decoder-only transformer architecture for unified learning across all input data. Research has demonstrated that this decoder-only framework offers enhanced generalization capabilities \cite{Wang2022whyDecoderOnly}, alongside a more streamlined structure, improved operational efficiency, and superior scalability.

\begin{figure}[h]
        \vspace{-1.5em}
    	\centering
    	\includegraphics[width=0.8\linewidth]{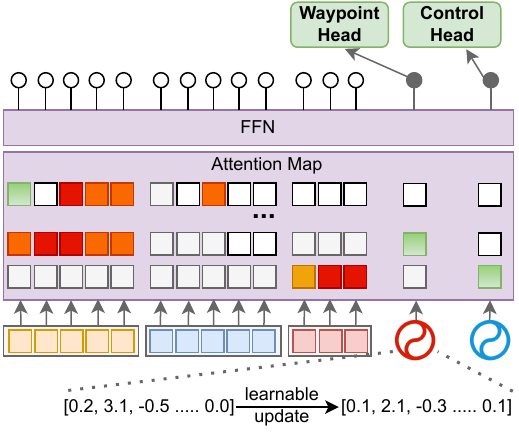}
        \caption{Decoder-only transformer with task guided learnable vector embedded.}
        \label{fig:decoder}
        \vspace{-1.5em}
    \end{figure}

    Contrary to the typical transformer-based architecture that interacts token features and then feeds them into the prediction head, our approach introduces a unique embedding vector for predictions. This vector is initialized using random values drawn from a Gaussian distribution, with feature representations for the prediction task being learned throughout training. This technique not only enhances the scalability of the decoder component but also leverages the attention mechanism to identify advantageous relationships between tokens at an early stage. The schematic diagram is shown in Fig. \ref{fig:decoder}.
    
    \subsubsection{Dynamic control}
    
    Contrary to TCP, which prioritizes the control volume predicted by the model and overlooks the control volume of the waypoint tracker, it's argued that the preferences of the underlying controller should be dynamically adapted based on the driving scenario.

    To accomplish this objective, a loss estimator is developed that utilizes the hidden features from the GRU within the waypoint head and the intermediate features from the Control head as its inputs, thereby modeling the training losses of both. The outcome of this process is leveraged as a measure of prediction confidence, which in turn is used to adjust the preference for the final control sequence. More specifically, this can be articulated as follows:
    \begin{equation}
        \label{equ:dynamicControl}
        \begin{matrix}
         \tilde{\mathcal L_c}  = Linear_1(X_c) \longrightarrow \mathcal L_c,\\
         \tilde{\mathcal L_w}  = Linear_2(X_w) \longrightarrow \mathcal L_w,\\
         Prefer = 1- \dfrac{\exp(k_c\tilde{\mathcal L_c})}{\exp(k_c\tilde{\mathcal L_c})+\exp(k_w\tilde{\mathcal L_w})}, 
        \end{matrix}
    \end{equation}
    where $\tilde{\mathcal L_w}$ and $\tilde{\mathcal L_c}$ are the predicted waypoint loss and control loss, which need to be as close as possible to the real loss $\mathcal L_w$ and $\mathcal L_c$. $k_c,k_w$ are coefficients used to make up for the difference in the numerical range of the two losses, which is determined based on experience.
    

%% file: section/4-Experiments.tex
\section{Experiments}
\input{table/main}
\subsection{Setup}
    \textbf{Evaluation metrics.} four metrics are used to evaluate the effectiveness of the methods: Driving Score (DS), Route Completion (RC), Number of parameters (Param), and Floating Point Operations (Flops). Among them, DS and RC are used to represent the driving effect, and Param and Flops are used to represent the efficiency of the neural network. 

    \textit{Driving Score (DS)} is the main metric of the Carla leaderboard\cite{Alexey2017DS}, serving as the product between the route completion and the infractions penalty. \textit{RC} is the percentage of the route distance completed by an agent. \textit{Param} refers to the total count of parameters that should be saved, which represents the size of the NN and usually determines the amount of storage space required to run the NN. \textit{Flops} is a measure of the computational cost required to perform a forward pass (inference) through the network. 

    \textbf{Dataset.} The CARLA\cite{Alexey2017CARLA} simulator is used for training and testing, specifically CARLA 0.9.10. 7 towns are used for training and Town05 for evaluation. Town05 is selected for evaluation due to the large diversity in drivable regions compared to other CARLA towns, e.g. multi-lane and single-lane roads, highways and exits, bridges and underpasses. To further enhance the assessment of driving safety, scenarios involving the sudden appearance of pedestrians and erratic vehicle behavior have been integrated into the simulation environment.  Two evaluation settings are considered: (1) Town05 Short: 32 short routes of 100-500m comprising 3 intersections each, (2) Town05 Long: 10 long routes of 1000-2000m comprising 10 intersections each. The weather condition is ClearNoon.

    \textbf{Training.} EfficientViT-m1 and m0 are used as visual backbone and load ImageNet\cite{Deng2009ImageNet} pre-trained weights. Other parts are initialized with Gaussian random numbers. EfficientFuser is trained for 60 epochs with a learning rate of 0.0005, and then trained for 60 epochs with a learning rate of 0.0001. The batch size is 256. Adam is used with weightDecay=1e-7. The learning rate is reduced to half every 30 epochs. Four parts of loss are set, namely speed loss $\mathcal L_s$, feature loss $\mathcal L_f$, waypoint loss $\mathcal L_w$, and control loss $\mathcal L_c$. $\mathcal L_f$ and $\mathcal L_s$ are the intermediate supervision to guide training, which is similar to TCP.
\subsection{Results}
    Table \ref{tab:main} presents the comparative analysis of EfficientFuser alongside other notable studies within the public Carla Leaderboard framework. EfficientFuser's performance is delineated in two distinct versions. The initial variant incorporates inputs from both the front view and the focus view (i.e., an enhanced frontal perspective), with an image resolution of 256x256. The 'Wider view' iteration maintains identical camera orientation but expands the image width to 768 pixels, thereby capturing a broader spectrum of information.

    EfficientFuser significantly reduces the size of closed-loop autonomous driving models derived from imitation learning to an unprecedented level. In comparison to the state-of-the-art lightweight approach, TCP, EfficientFuser's model size is merely 37.6\% as large, and its computational demand is just 8.5\% of TCP's. Despite its considerably smaller neural network (NN) size, EfficientFuser only experiences a marginal decrease of 0.4\% in Driving Score (DS) in the Town05 Short scenario. When compared with CIL under similar parameters and computational loads, EfficientFuser substantially outperforms it, showing a remarkable 73-point advantage in DS. 

    As a cutting-edge solution for accessible closed-loop autonomous driving, InterFuser achieves a similar Route Completion (RC) to EfficientFuser but improves DS by 6.3 points. This improvement, however, comes at the expense of the number of parameters 8.4$\times$ and computational requirements 31.6$\times$. Based on detection outcomes, these models often resort to a cautious approach, suggesting that the vehicle would proceed at a slow pace when it has been stationary for an extended period without any obstacles in its path. With its enhanced safety driving strategy, InterFuser would even recognize the red light at the next unreachable intersection far away and come to a halt — a behavior not aligned with typical human driving patterns, which EfficientFuser does not exhibit.

    To highlight the security efficacy of EfficientFuser,  the penalties incurred for infractions by various methods are reported on Town05 Short. EfficientFuser significantly surpasses the security benchmarks set by both TCP and Transfuser, and it only marginally falls short of Interfuser, which prioritizes security at its core. 
    \begin{figure}[h]
        \vspace{-1.5em}
    	\centering
    	\includegraphics[width=\linewidth]{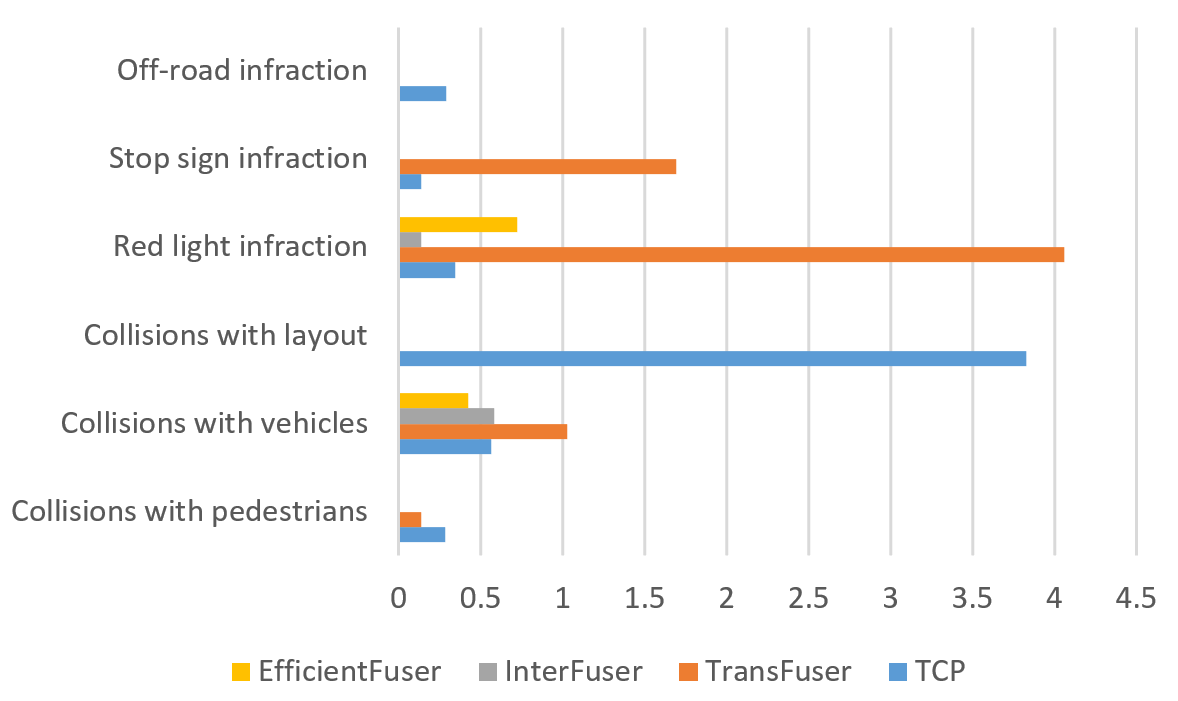}
        \caption{The penalties incurred for infractions.}
        \label{fig:infractions}
        \vspace{-1.5em}
    \end{figure}
    EfficientFuser employs a novel approach by introducing a learnable embedding vector for predictions. This methodology diverges from traditional practices that rely on feeding sensor token information directly into the prediction head. By adopting this design, EfficientFuser begins to explore the correlation between the prediction target and the input data at an early stage, thereby generating task-specific representations. To illustrate the impact of this approach, two representative attention maps are visualized. The red lines divide the tokens from different information sources. Taking the abscissa as an example, from left to right they are prediction tokens, measurement tokens, side-view tokens and main-view tokens.
    
    \begin{figure}[htbp]
    	\centering
    	\begin{minipage}{0.45\linewidth}
    		\centering
    		\includegraphics[width=0.85\linewidth]{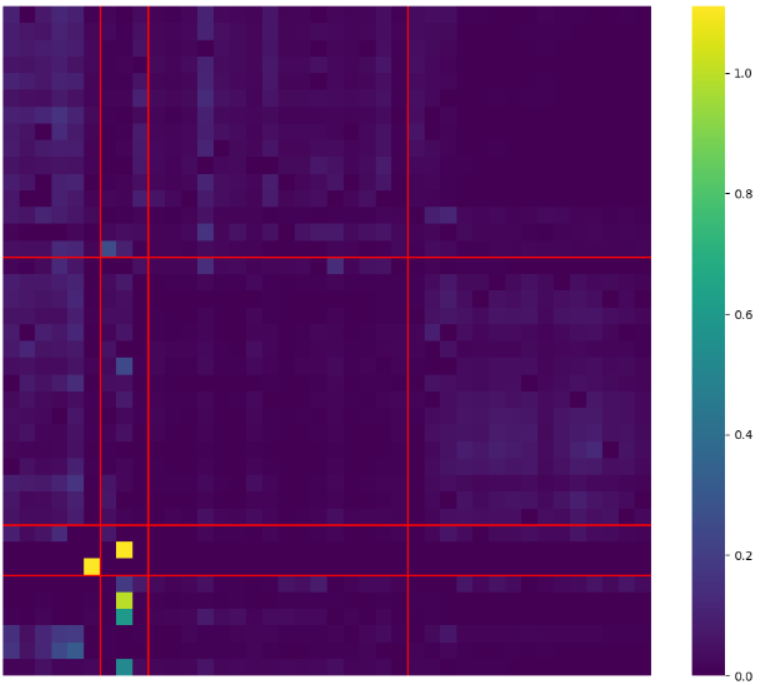}
    		\subcaption{\textbf{Layer0 Head3.}
                    Prediction tokens focus on the measurement tokens.}
    		\label{fig:attentionmap1}
    	\end{minipage}
            \hspace{1em}
    	\begin{minipage}{0.45\linewidth}
    		\centering
    		\includegraphics[width=0.85\linewidth]{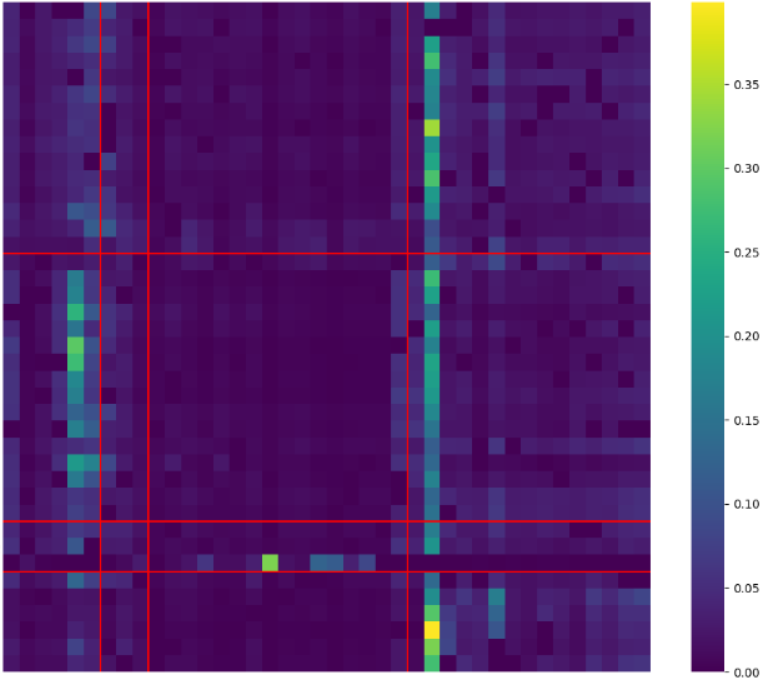}
    		\subcaption{\textbf{Layer4 Head2.} Information from a specific patch in the main view is accentuated.}
    		\label{fig:attentionmap2}
    	\end{minipage}
     
        \caption{Attention Maps in Decoder.}
        \label{fig:attentionmap}
        \vspace{-1.5em}
    \end{figure}

\subsection{Ablation Study}
    To thoroughly investigate the system architecture and evaluate the effectiveness of our proposed method, we have carried out a series of ablation studies. Initially, the impact of utilizing image backbones of varying sizes and the performance enhancement brought about by the cross-attention fuser were assessed. Subsequently, we explored the effects of altering the decoder layer depths and implementing learnable vector embeddings. Lastly, we compared the dynamic control adjustment with the static allocation approach employed in TCP.
    
    The outcomes of these experiments were all gathered on Town05 Short, providing a comprehensive analysis of each component's contribution to the overall performance.
    \subsubsection{Image Backbone}
        Fusion with cross attention is an efficient method, resulting in only 5.7\% additional parameters and 7.2\% computation in EfficientFuser.Despite its minimal computational requirement, Cross Attention plays a pivotal role in EfficientFuser. The absence of this multi-level image information interaction significantly impacts the system's efficacy, leading to a notable reduction of 14.9 points in the driving score. This underlines the critical importance of Cross Attention in facilitating effective feature fusion and enhancing the model's ability to make informed decisions.
    
        Intuitively, one might assume that a larger image backbone would yield superior results, given its enhanced capabilities for feature extraction. Nevertheless, as demonstrated in Tab.\ref{tab:imageBackbone}, an increase in backbone size actually detracts from the model's performance. This counterintuitive outcome may be attributed to the complexity of the training tasks, which hinders the full training potential of larger backbones.
        \input{table/imageBackbone}

    \subsubsection{Decoder Layer}
        In evaluating our decoder layer, we established two sets of comparative analyses. The initial set investigates the impact of varying the depth of the decoder layer. Following this, we examine our proposed method of employing learnable vectors as prediction tokens. For comparative purposes, we conducted an additional experiment wherein, rather than utilizing the learnable vector and integrating it into the token dimension, we averaged the sensor features and sent them into the prediction head to observe the effect on predictions. The outcomes of these investigations are presented in Tab. \ref{tab:decoderLayers}.
        \input{table/decoderLayers}
        Although it introduces a minor computational overhead, the learnable vector approach utilized by EfficientFuser significantly enhances the driving performance. Concerning the depth of the decoder layer, it is evident that increasing the depth does not necessarily improve performance. While the RC of the vehicle may increase, the DS substantially decreases, indicating that the vehicle starts to overlook safety-related information. This phenomenon can also be observed in the visual attention maps; beginning from the seventh layer, the focus shifts away from the prediction token towards reinforcing certain measurement token information.
    \subsubsection{Dynamic Control}
        EfficientFuser adopts the hybrid control utilized by TCP and incorporates a dynamic allocation strategy. To validate the efficacy of the dynamic adjustment method, the experiments about the TCP settings and dynamic setting are introduced. The outcomes in Tab. \ref{tab:control} affirm the effectiveness of the dynamic allocation component, enhancing the DS and RC.
        \input{table/control}
    

%% file: table/main.tex
\begin{table*}[t]
  \centering
  \renewcommand{\arraystretch}{1.1} 
  \caption{\textbf{Driving performance.} We report the mean deviation over 3 runs of each method. $\downarrow$ indicates that the smaller the value, the better. $\uparrow$ indicates that the bigger the value, the performance better.}
  \label{tab:main}
  \begin{tabular}{c|cc|cc|cc|cc}
    \hhline
    \multirow{2}[1]{*}{\textbf{Method}} & \multicolumn{2}{c|}{\textbf{Sensor}} & \multirow{2}[1]{*}{\textbf{Param(M)$\downarrow$}} & \multirow{2}[1]{*}{\textbf{Flops(G)$\downarrow$}} & \multicolumn{2}{c|}{\textbf{Town05 Short}} & \multicolumn{2}{c}{\textbf{Town05 Long}} \\

    & Camera & Lidar & & & DS$\uparrow$ & RC$\uparrow$ & DS$\uparrow$ & RC$\uparrow$ \\
    \hline

    CIL & $\checkmark$ & $\times$ & 12.7 & 2.4 & 7.5& 13.4 & 3.7 & 7.2 \\
    LBC & $\checkmark$ & $\times$ & 13.0 & 3.16 & 31.0 & 55.0 & 7.1 & 32.1 \\
    AIM\cite{Prakash2021transfuser1} & $\checkmark$ & $\times$ & 21.5 & 4.8 & 49.0 & 81.1 & 26.5 & 60.7 \\
    Late Fusion & $\checkmark$ & $\checkmark$ & 32.6 & 7.1 & 51.6 & 81.1 & 26.5 & 60.7 \\
    Transfuser & $\checkmark$ & $\checkmark$ & 66.1 & 11.4 & 54.5 & 78.4 & 33.1 & 56.3 \\
    TCP & $\checkmark$ & $\times$ & 25.8 & 17.1 & 80.9 & 96.1 & 45.1 & 71.5 \\
    InterFuser & $\checkmark$ & $\checkmark$ & 82.2 & 46.1 & 86.9 & 92.7 & 33.6 & 56.7\\

    \hline
    \textbf{EfficientFuser}  & $\checkmark$ & $\times$ & 9.7 & 0.51 & 75.8 & 85.2 & 40.6 & 56.7\\
    \textbf{EfficientFuser(wider view)}  & $\checkmark$ & $\times$ & 9.7 & 1.46 & 80.5 & 92.2 & 42.7 & 59.1\\
    \hhline
  \end{tabular}
  \vspace{-1.5em}
\end{table*}

%% file: table/imageBackbone.tex
\begin{table}[!htb]
  \centering
  \renewcommand{\arraystretch}{1.1} 
  \caption{Ablation on the image backbone}
  \label{tab:imageBackbone}
  \begin{tabular}{c|cc|cc}
    \hhline
    \textbf{Backbone}   &  \textbf{Cross Attn}   & \textbf{Param(M)}       & \textbf{DS} & \textbf{RC}\\ 
    \hline
    $EfficientViT_{m1}$ &  \multirow{2}[1]{*}{No}&  \multirow{2}[1]{*}{9.2}  &\multirow{2}[1]{*}{60.9} & \multirow{2}[1]{*}{76.3} \\
    $EfficientViT_{m0}$ & &  & &\\
    \hline

    $EfficientViT_{m1}$ &  \multirow{2}[1]{*}{Yes}&  \multirow{2}[1]{*}{9.7}  &\multirow{2}[1]{*}{75.8} & \multirow{2}[1]{*}{85.2} \\
    $EfficientViT_{m0}$ & &  & &\\
    \hline
    
    $EfficientViT_{m5}$ &  \multirow{2}[1]{*}{Yes}&  \multirow{2}[1]{*}{37.5} & \multirow{2}[1]{*}{65.3} & \multirow{2}[1]{*}{82.0} \\
    $EfficientViT_{m4}$ & &  & &\\
    \hhline
  \end{tabular}
  \vspace{-1em}
\end{table}

%% file: table/decoderLayers.tex
\begin{table}[!htb]
\vspace{-1em}
  \centering
  \renewcommand{\arraystretch}{1.1} 
  \caption{Ablation on decoder layers}
  \label{tab:decoderLayers}
  \begin{tabular}{c|ccc|cc}
    \hhline
    \multirow{2}[1]{*}{\textbf{Depth}}   &   \textbf{Learnable}   & \multirow{2}[1]{*}{\textbf{Param(M)}}    & \multirow{2}[1]{*}{\textbf{Flops(G)}}   & \multirow{2}[1]{*}{\textbf{DS}} & \multirow{2}[1]{*}{\textbf{RC}}\\
    & \textbf{Vector} & & & & \\
    \hline

    1 & Yes & 6.6 & 0.39 &  51.4 & 62.4  \\
    4 & Yes & 8.0 & 0.44 &  71.2 &   80.4\\
    8 & No & 9.7 & 0.51 & 70.0  &  82.7 \\
    \textbf{8} & \textbf{Yes} & \textbf{9.7} & \textbf{0.51} & \textbf{75.8} & \textbf{85.2} \\
    12 & Yes  &11.5 & 0.59 &  67.0 & 90.2  \\
    \hhline
  \end{tabular}
\end{table}

%% file: table/control.tex
\begin{table}[!htb]
\vspace{-1em}
  \centering
  \renewcommand{\arraystretch}{1.1} 
  \caption{Ablation on the control strategy}
  \label{tab:control}
  \begin{tabular}{c|cc}
    \hhline
    \textbf{Method}  & \textbf{DS} & \textbf{RC}\\ 
    \hline
    Same with TCP & 72.4 & 81.0\\
    With dynamic adjustment & 75.8 & 85.2 \\
    \hhline
  \end{tabular}
\end{table}

%% file: section/5-Conclusion.tex
\section{Conclusion}
    
    EfficientFuser markedly diminishes both the size and computational demands of neural networks by incorporating a compact and efficient vision transformer as its visual backbone, coupled with the use of cross-attention mechanisms for the integration of information. The adoption of a decoder-only transformer architecture alongside learnable prediction vectors ensures that the system maintains commendable performance despite its reduced scale. Furthermore, alterations to the fundamental controller not only boost the driving efficiency but also significantly augment the safety features of the system. As the smallest closed-loop autonomous driving NN available to date, EfficientFuser stands out for its exceptional performance. However, similar to TCP, EfficientFuser not only learns from the results but also from intermediate features, which makes it suffer from some performance degradation when sim-to-real.